\documentclass[letterpaper, 10 pt, conference]{ieeeconf}  
\IEEEoverridecommandlockouts                              

\usepackage{amssymb}  

\usepackage{float}
\usepackage[normalem]{ulem}
\usepackage{amsmath}
\usepackage[noadjust]{cite}
\usepackage{siunitx}
\usepackage{booktabs}
\usepackage[flushleft]{threeparttable}

\makeatletter
\let\NAT@parse\undefined
\makeatother
\usepackage[colorlinks,bookmarksopen,bookmarksnumbered,citecolor=red,urlcolor=red]{hyperref}
\usepackage[colorlinks,bookmarksopen,bookmarksnumbered,citecolor=red,urlcolor=red]{hyperref}
\usepackage[]{cleveref}

\title{\LARGE \bf
FruitTouch: A Perceptive Gripper for Gentle and Scalable\\ Fruit Harvesting
}

\usepackage[colorinlistoftodos]{todonotes}

\usepackage{comment}

\begin{document}
\author{Ruohan Zhang$^{1}$,
        Mohammad Amin Mirzaee$^{1}$,
        and Wenzhen Yuan$^{1}$
\thanks{$^{1}$Ruohan Zhang, Mohammad Amin Mirzaee, and Wenzhen Yuan are with University of Illinois at Urbana-Champaign, Champaign, IL, USA 
        {\tt\small \{rz21,mirzaee2,yuanwz\}@illinois.edu}}%
        }


\maketitle

\thispagestyle{empty}
\pagestyle{empty}


\begin{abstract}
The automation of fruit harvesting has gained increasing significance in response to rising labor shortages. A sensorized gripper is a key component of this process, which must be compact enough for confined spaces, able to stably grasp diverse fruits, and provide reliable feedback on fruit conditions for efficient harvesting.
To address this need, we propose FruitTouch, a compact gripper that integrates high-resolution, vision-based tactile sensing through an optimized optical design. This configuration accommodates a wide range of fruit sizes while maintaining low cost and mechanical simplicity. Tactile images captured by an embedded camera provide rich information for real-time force estimation, slip detection, and softness prediction. We validate the gripper in real-world fruit harvesting experiments, demonstrating robust grasp stability and effective damage prevention. The hardware design files and simulation environment are open-sourced at project website: (\url{https://fruittouch-dev.github.io/fruittouch/})


\end{abstract}

\section{introduction}

Agricultural robotic systems for harvesting have been a focus of the community due to the lack of human labor. When designing automated harvesting systems, the choice of the end effector is critical since it is directly responsible for handling delicate fruits, preventing slips, inferring fruit firmness, and preserving the quality of the fruits by avoiding crushing or rubbing during the harvest process \cite{HUSSEIN20201}. Existing agricultural end-effectors commonly rely on visual feedback for fruit detection \cite{koe2025precisionharvestingclutteredenvironments} and mechanical designs such as suction cups \cite{8842591, reddy2013review}, scissor cutters \cite{Zhang2025TomatoHarvest}, or multi-finger grippers \cite{https://doi.org/10.1002/rob.21715, GAO2022106879} to enhance harvesting success rate. While effective under ideal conditions, these solutions face several limitations. Vision-based feedback can be unreliable when the fruit is occluded by foliage or when lighting conditions vary. Purely mechanical end-effectors 
often lack robustness to environmental variability, including fruit size variation, clustered growth, or surface wetness.
While some of these systems later augment vision with low-resolution or binary fingertip contact sensors to signal contact onset or threshold events \cite{s22155483, Zhang2025SlipControl}, such signals provide little spatial or directional information. As a result, they cannot recover the rich contact state and fruit material properties for achieving reliable grasping. 


On the contrary, human pickers excel at this task by integrating tactile and force cues to detect contact state and assess ripeness during manipulation. Recent advances in high-resolution tactile sensing, such as GelSight \cite{yuan2017gelsight,mirzaee2025gelbelt,gelwedge, zhang2025pneugelsightsoftroboticvisionbased} technology, offer a promising path toward overcoming these limitations. GelSight sensors capture detailed surface geometry and deformation patterns, enabling accurate estimation of contact forces \cite{zhang2022deltactvisionbasedtactilesensor,zhang2019effective}, detection of slip \cite{hu2024learningdetectsliptactile, taylor2021gelslim3}, and assessment of surface compliance \cite{TomatoFirmnessVisionTactile
}. These capabilities have driven successful applications of GelSight in various robotic manipulation tasks; however, its integration into agricultural grippers remains limited.

\begin{figure}
    \centering
    \includegraphics[width=\linewidth]{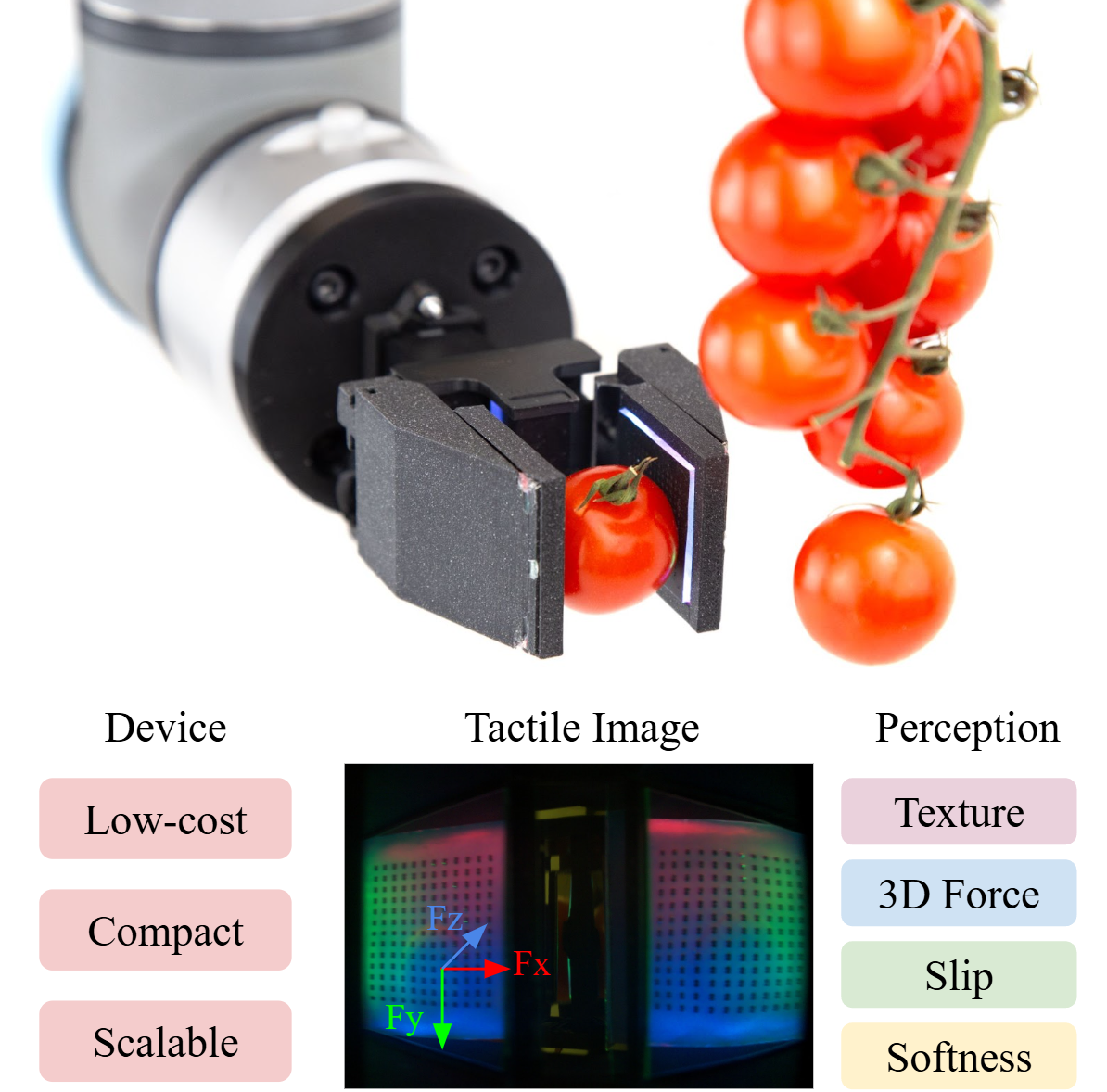}
    \vspace{-15pt}
    \caption{Demonstration of the FruitTouch gripper harvesting cherry tomatoes. We designed the gripper for compactness, low cost, and scalability, while the perception system measures high-resolution contact geometry, 3D force, slip, and object softness. The design integrates mechanical efficiency with rich tactile sensing to enable reliable and efficient fruit harvesting. 
    }
    \label{fig:teaser}
    \vspace{-18pt}
\end{figure}

In this work, we present FruitTouch, a compact and lightweight agricultural end-effector that integrates high-resolution tactile sensing to enable robust perception and harvesting in real-world scenarios. Figure \ref{fig:teaser} provides an overview of the FruitTouch gripper and its application in harvesting.
The design adopts a parallel-jaw mechanism with a single camera to sensorize both gripping surfaces, reducing cost and complexity for fabrication. Its thin, wedge-shaped fingers allow the gripper to reach into dense foliage and clutter, which is critical for practical field deployment.
Moreover, the scalable mechanical and optical design opens up the possibility of accommodating fruit sizes ranging from those smaller than cherry tomatoes ($\sim$28 mm diameter) to large fruits like apples ($\sim$75 mm diameter).
To summarize, our proposed FruitTouch gripper can:


\begin{itemize}
    \item Reconstruct fruit surface texture with high precision,  
    \item Detect and prevent fruit crushing through accurate 3D force measurement,  
    \item Identify and respond to slip events during grasping, and  
    \item Classify fruit softness as a proxy for ripeness.  
\end{itemize}



We use cherry tomatoes and strawberries as representative soft fruits for discussion—vine-grown produce that often grows in clusters, with relatively small diameters (e.g., cherry tomatoes averaging approximately 28.3 mm \cite{GAO2022106879}) and delicate surfaces prone to bruising. This presents significant challenges for common end-effectors due to the need for gentle yet secure handling in cluttered and constrained environments. By integrating high-resolution tactile sensing, a compact and scalable mechanical and optical design, and adaptability for agricultural settings, our system demonstrates strong performance in these demanding conditions, achieving high precision and operational efficiency during harvesting. We believe this work can contribute toward enabling scalable, automated harvesting solutions that reduce harvest losses, increase harvesting efficiency, and lessen reliance on human labor in the foreseeable future.


\section{Related Work}
\subsection{Agricultural End-Effectors for Harvesting}
According to the detachment method, agricultural end-effectors can be broadly classified into three categories: cutting-based, suction-based grasping, and mechanical clamping approaches \cite{https://doi.org/10.1002/rob.70021, agronomy15112650}. Cutting-based methods detach fruit from the stem and are well suited to crops with long, accessible stems, such as strawberries and tomatoes \cite{HAYASHI2010160, 9382670}. Suction-based grasping relies on negative pressure to attach to the fruit surface, enabling low-damage harvesting for lightweight fruits with smooth surfaces. In contrast, mechanical clamping mimics human harvesting behavior by detaching fruit through direct surface contact and the application of pulling, twisting, or bending forces, offering strong potential for generalization across diverse crop types \cite{WILLIAMS2019140}. However, such direct contact can increase the risk of fruit bruising during interaction, which limits the applicability of purely mechanical clamping approaches in damage-sensitive harvesting tasks.

While detachment strategy determines how force is applied to separate fruit from the plant, harvesting performance is also strongly influenced by how contact forces are transmitted and regulated at the gripper–fruit interface. From this perspective, end-effectors can also be categorized by finger material and compliance, ranging from fully rigid designs, to hybrid configurations with rigid backbones and compliant elastomeric contact layers, to fully soft grippers. Fully rigid grippers are commonly used for harvesting rigid crops such as pumpkins and watermelons but risk causing surface damage \cite{ROSHANIANFARD2020105503}. Fully soft grippers provide higher compliance but often suffer from limited load capacity and increased fabrication and control complexity, making them better suited to less demanding harvesting scenarios \cite{s23187905}. Hybrid designs seek to balance these trade-offs by combining structural reliability with compliant surface contact.

Situated at the intersection of these two taxonomies, the proposed FruitTouch gripper is best characterized as a hybrid rigid clamping end-effector. While such designs offer robust grasping and damage mitigation, their effectiveness depends on sensing and exploiting tactile interaction during grasping, as contact forces must be carefully regulated to accommodate variations in fruit size, stiffness, and surface condition. Vision-based tactile sensing provides a practical and effective means to capture this information, motivating its integration in our system.

\subsection{Perception Using Vision-based Tactile Sensors}
Vision-based tactile sensors, particularly GelSight \cite{firstgelsight}, have shown great potential for capturing detailed contact geometry and providing rich feedback for robotic manipulation tasks. Such sensors can enable accurate force estimation, slip detection, and object softness detection, which make it suitable for integration into agriculture end-effectors.

For example, \cite{yuan2017gelsight} demonstrated that the relationship between indentation volume and normal force is approximately linear. Building on this insight, subsequent works \cite{ma2019dense,10758225} employed finite element methods (FEM) for more refined force estimation. Shear force has been estimated by tracking the displacement of surface markers  \cite{yuan2017gelsight,zhang2019effective}, which have also been leveraged for slip detection: average marker displacement has been used for small objects \cite{yuan2015measurement}, while entropy-based measures have been applied to larger contact areas \cite{hu2024learningdetectsliptactile}. Object softness prediction has likewise been an active area of research, with early approaches modeling softness as a linear function of image brightness \cite{7759057}, and more recent studies incorporating analytical models to better interpret tactile signals and improve compliance estimation \cite{burgess2025learningobjectcomplianceyoungs}.

Despite their potential for enhancing perception, existing vision-based tactile systems often suffer from bulky form factors, high cost, limited capability, or poor generalizability, making them difficult to apply directly to harvesting. For example, \cite{https://doi.org/10.1002/adrr.202500020}
presented a GelSight-equipped 5-DoF gripper for in-hand manipulation, but its complex multi-DOF structure runs counter to the simplicity and scalability required in agricultural applications. Likewise, \cite{TomatoFirmnessVisionTactile} used GelSight as a standalone post-harvest firmness tester rather than integrating it into the picking process, while \cite{VISENTIN2023108202} developed a strawberry-harvesting end-effector that, although capable of force feedback, remained bulky and lacked both slip detection and ripeness estimation.
 

In contrast, our proposed gripper is sensorized, compact, and low-cost, providing integrated perception of force, slip, and ripeness within a single harvest-ready design. Unlike prior GelSight-based sensors for agriculture, it employs a single camera shared across both fingers through a common optical path. This approach reduces cost while maintaining sensing effectiveness under varying gripper configurations, enabling practical deployment in dynamic harvesting environments.

\begin{figure*}
    \centering
    \includegraphics[width=\linewidth]{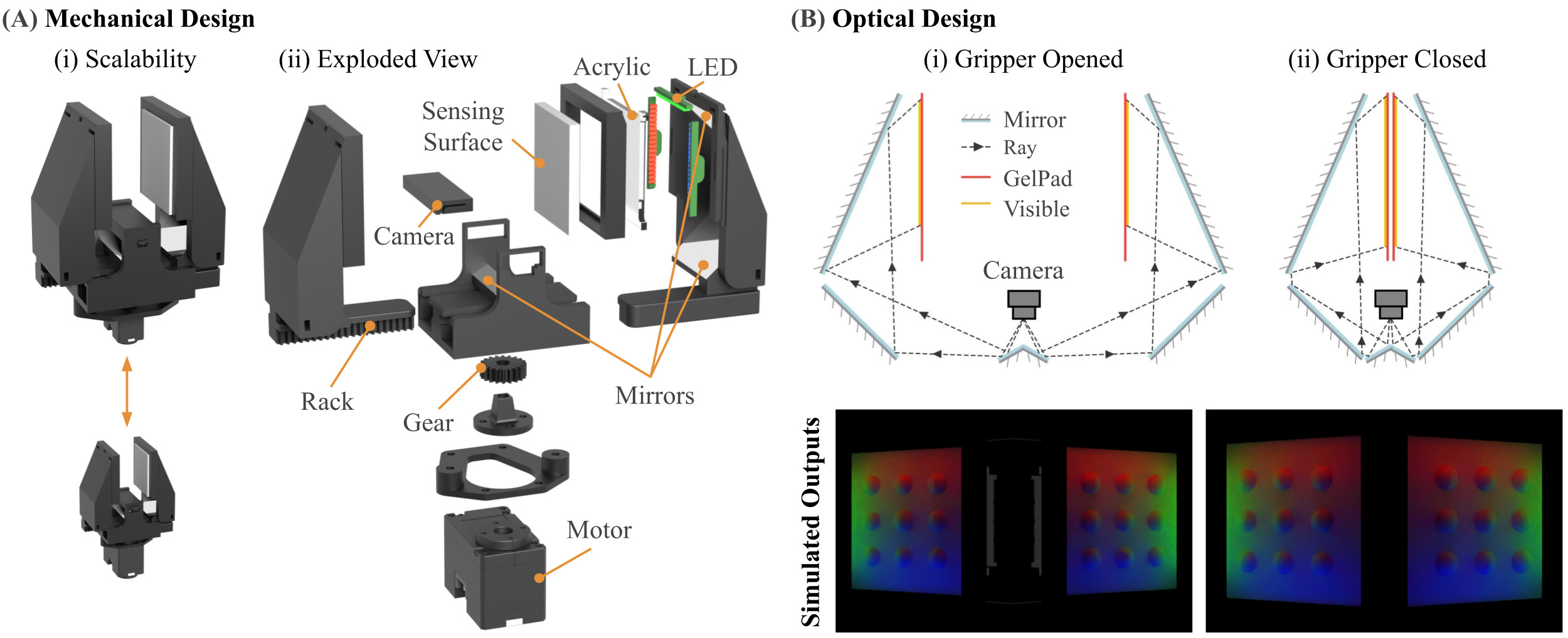}
    
    \caption{\textbf{Mechanical and Optical Design of FruitTouch Gripper}.
        \textbf{(A)} Mechanical Design. The gripper components are designed for scalability, enabling harvesting of fruits with varying sizes. We use gear-and-rack mechanism to provide actuation. Each finger consists of a soft silicone sensing surface supported by a transparent acrylic sheet, with three LED strips ensuring uniform illumination. A centrally mounted camera, in combination with mirrors, provides comprehensive coverage of both sensing surfaces.
        \textbf{(B)} Sensor Optical Design. The mirror configuration is optimized to maximize the sensing area while maintaining low distortion across different finger distances. Simulated outputs are shown for both the open and closed states of the gripper.
        }
    \vspace{-18pt}
    \label{fig:design}
\end{figure*}

\section{Sensor-Integrated Gripper Design}

To achieve a practical balance between simplicity, robustness, and sensing capability,
we adopt a parallel-jaw gripper design to reduces mechanical complexity and minimizes interference with both the optical system and the surrounding environment. 
This also mimics the natural two-finger picking strategy, with integrated tactile sensing on both fingers. The following subsections present the mechanical specifications of the gripper in \Cref{sec:mechD} and the optical design in \Cref{sec:optD}.

\subsection{Mechanical Design}
\label{sec:mechD}
We placed the racks and gear at the base of the gripper to actuate the fingers and to minimize interference with the optical subsystem. Racks are attached to the fingers and coupled with a gear connected to a DYNAMIXEL XC330-M288-T motor, as shown in \Cref{fig:design}A. 

The smallest version of the FruitTouch gripper was designed to handle small fruits such as cherry tomatoes. The sensing surface, based on the mechanical configuration, employed a target gelpad size of $30 \text{ mm}$ and a stroke of approximately $40 \text{ mm}$.  Excluding the motor, the closed gripper measures $35 \times 40 \times 66 \text{ mm}^3$, and the fingers can open up to $40 \text{ mm}$, providing sufficient range to grasp small fruits including grapes and strawberries. The gripper design is geometrically scalable: both the mechanical structure and mirror-based optical layout can be uniformly scaled while preserving kinematic and ray-tracing relationships. In practice, the minimum achievable scale is only constrained by the physical dimensions of the camera.


\subsection{Optical Design}
\label{sec:optD}

The optical subsystem consists of two components: the tactile sensing surfaces integrated into the gripper fingers, and a single camera positioned between them. Intermediate mirrors split the camera’s field of view into two halves, each corresponding to a fingertip, while LEDs positioned near the gelpads enhance illumination for tactile signal acquisition.

To determine the optimal configuration of the mirrors and sensing surfaces, we simulated the sensor in two stages.
First, using the 2D Ray Optics Simulation on phydemo.app, we coarse-aligned three mirrors relative to the camera for each finger.
Our goal was to orient the optical components such that the incident rays strike the sensing surface as orthogonal as possible, since this minimizes perspective distortion in each frame and maintains similar tactile signals in different finger distances. The result of this stage is shown in \Cref{fig:design}B for the open and closed configuration of the finger. We observed that both cases yield similar perspectives with acceptable levels of distortion. Moreover, the optical configuration is fully scalable while preserving the same coverage ratio.

We used the geometrical information of this step to run a more detailed optical simulation in Blender to fine-tune the mirror configuration and optimize the light locations.
We adopted the method from \cite{Agarwal_2025} to optimize the location of the LEDs, obtaining an improved color matching to surface normal (RGB2Norm Metric). This yields a high-quality optical system, as evidenced by the minimally distorted spherical indentations in \Cref{fig:design}B, which in turn enhances the accuracy of geometry calibration and reconstruction.
We also simulated a scaled version of our gripper, where the tactile signals remained consistent and the only variation was in the pixel-to-millimeter ratio.


\section{Method of Perception}


\begin{figure}
    \centering
    \includegraphics[width=\linewidth]{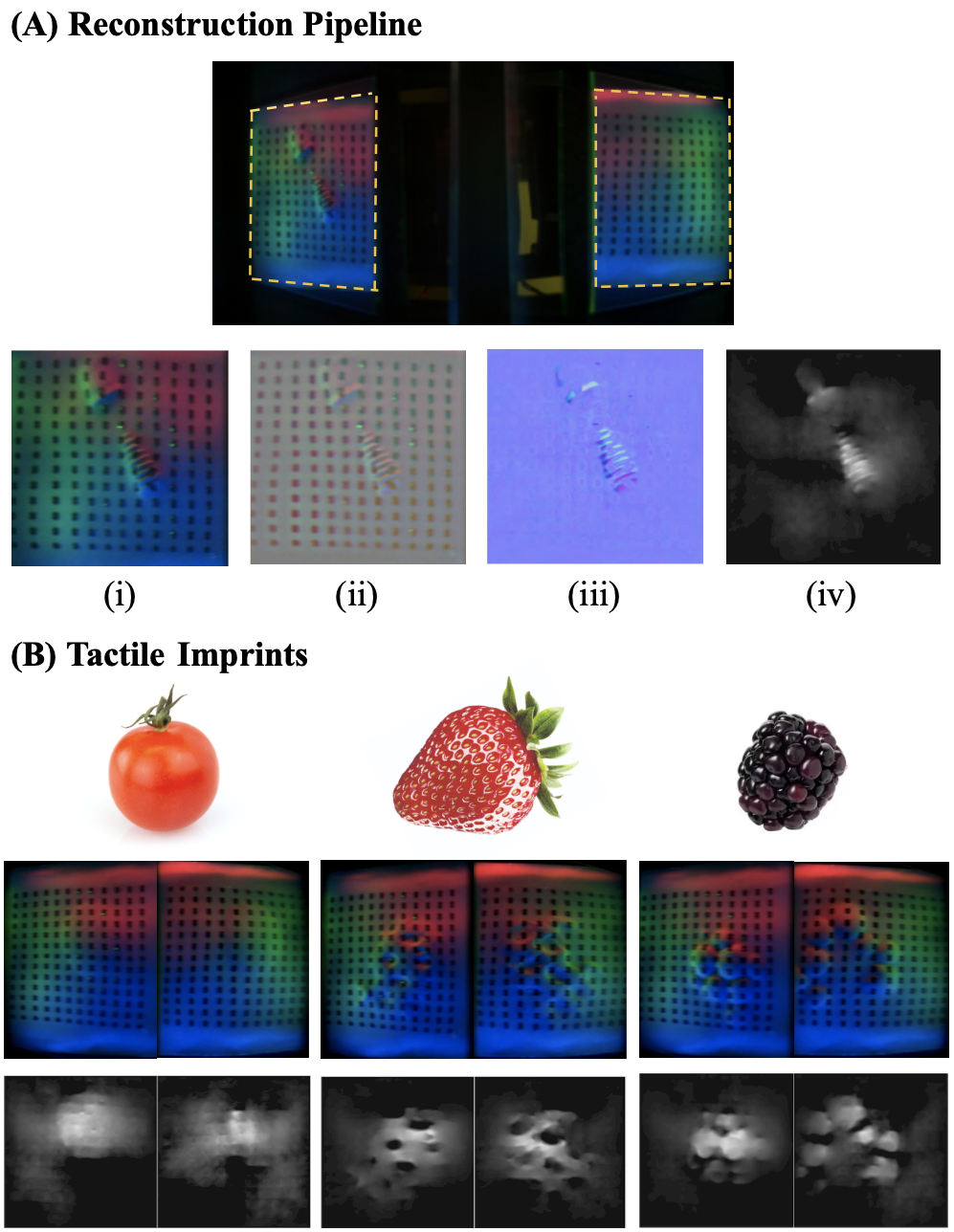}
    \vspace{-15pt}
    \caption{\textbf{Tactile Sensing Pipeline}. \textbf{(A)} Contact-geometry reconstruction pipeline. \textbf{Top:} Raw camera output that contains reading from both fingers. \textbf{(i)} Unwarped contact frames on one sensor. \textbf{(ii)} Background-subtracted difference images. \textbf{(iii)} Estimated surface normals. \textbf{(iv)} Reconstructed 3D shape (shown in the form of a heightmap). \textbf{(B)} Example tactile images and the reconstructed shape for a cherry tomato, a strawberry, and a blackberry.}
    \label{fig:calibration}
    \vspace{-18pt}
\end{figure}

\subsection{Tactile Geometry Reconstruction}



Tactile geometry reconstruction is a core capability of vision-based tactile sensing and it enables downstream tasks such as fruit-type classification, bruise/defect detection, and contact-state reasoning. The idea was first demonstrated by \cite{firstgelsight}, which used colored illumination and an internal camera to map RGB intensities to local surface normals via photometric stereo. From these normals, the contact geometry can be represented directly as a dense normal field or integrated to yield a height map/mesh that captures fine contact textures.

In our work, geometry reconstruction is achieved by first calibrating the sensor through indentations with steel balls of known radius (5 mm). The raw tactile image is rectified into a rectangular frame, as shown in \Cref{fig:calibration}A, and the background image is subtracted to enhance the contact signal. The resulting contact images are paired with manually labeled surface normals to build a calibration dataset. Using this dataset, we train a MLP to learn the mapping from RGB pixel intensities to surface normals. At inference time, the predicted normals are numerically integrated to recover the height map of the indentation, thereby reconstructing the local contact geometry. The complete pipeline is illustrated in \Cref{fig:calibration}A.

\subsection{Three-dimensional Force Estimation}

Three-dimensional force estimation, including both normal and shear components, is essential for closed-loop control in fruit harvesting. For example, normal force estimation helps maintain appropriate grasp strength, while shear force prediction assists in preventing slip. To achieve this, we leverage FruitTouch’s physical properties with a lightweight, data-driven model that fuses actuator signals and visual cues from the gel surface for real-time inference.

In our parallel-jaw gripper setting, normal forces at the fingertips generate opposing contact loads that are transmitted through the racks and gear to the motor,
increasing actuator torque and proportionally increasing motor current under fixed gearing.
We therefore fit a simple linear model that maps the current \(I\) to the normal force \(F_n\).

For shear force estimation, we infer both magnitude and direction by visually tracking the displacement of etched surface markers on the gel. Following the magnetic-field analogy and Helmholtz–Hodge decomposition described in \cite{zhang2019effective}, the displacement field is separated into curl-free and divergence-free components, which correspond to translational and rotational shear patterns, respectively. A feature vector is then constructed by applying low-order polynomial expansions to these components.

To be specific, the 2D marker displacement field $\mathbf{V}$ is decomposed into three orthogonal components:

\begin{equation}
\mathbf{V} = \mathbf{P} + \mathbf{S} + \mathbf{H}
\end{equation}
\vspace{-10pt}

Where $\mathbf{P}$ denotes the curl-free (irrotational) component with $\nabla \times \mathbf{P} = 0$, $\mathbf{S}$ is the divergence-free (solenoidal) component with $\nabla \cdot \mathbf{S} = 0$, and $\mathbf{H}$ is the harmonic component satisfying both $\nabla \cdot \mathbf{H} = 0$ and $\nabla \times \mathbf{H} = 0$. Denoting the interpolated marker components as $\mathbf{p}, \mathbf{s}, \mathbf{h}$ and their respective scalar components using subscripts (e.g., $\mathbf{p}_x, \mathbf{p}_y$), the shear deformation feature is defined as:

\vspace{-5pt}
\begin{equation}
    \begin{aligned}
        &\mathbf{x}_{\text{shear}} = 
        \Big[ \mathbf{v}_x, \mathbf{v}_y, 
        \mathbf{p}_x^*, \mathbf{p}_y^*, \mathbf{s}_x^*, \mathbf{s}_y^* \Big]^T, \\
        &\mathbf{p}_x^* = \big[ \mathbf{p}_x, \mathbf{p}_x^2 \big]^T \quad
        \mathbf{p}_y^* = \big[ \mathbf{p}_y, \mathbf{p}_y^2 \big]^T  \\
        &\mathbf{s}_x^* = \big[ \mathbf{s}_x, \mathbf{s}_x^2 \big]^T \quad \ \ \,
        \mathbf{s}_y^* = \big[ \mathbf{s}_y, \mathbf{s}_y^2 \big]^T 
    \end{aligned}
\end{equation}
\vspace{-5pt}

and the total shear force prediction can be written as
\vspace{-5pt}
\begin{equation}
    \mathbf{F_{shear}} =
    \begin{bmatrix}
        F_x \\[2pt]
        F_y \\[2pt]
    \end{bmatrix}
    =
    \begin{bmatrix}
        \mathbf{w}_x^T \\
        \mathbf{w}_y^T \\
    \end{bmatrix}
    \begin{bmatrix}
        \mathbf{x}_{\text{shear}} \\
    \end{bmatrix}
    +
    \begin{bmatrix}
        b_x \\[2pt]
        b_y \\[2pt]
    \end{bmatrix}
\end{equation}
\vspace{-5pt}


\subsection{Real-Time Slip Detection}
In agricultural harvesting, slip refers to unintended relative motion between the end effector and the fruit, often leading to damage, grasp failure, or reduced efficiency. Achieving stable agricultural harvesting, therefore, requires critical capabilities in slip detection. Inspired by \cite{dong2017improved}, our approach detects slip by directly comparing the motion of a grasped object with the displacement of surface markers, a method made possible by high-resolution contact sensing.

To estimate the object's motion, we first segment the contact region by thresholding the reconstructed height map. The average velocity of this region's center point serves as a proxy for the object's velocity. We then compute the marker velocities within the same contact region and evaluate the difference between the object and marker velocities. A slip event is reported when this difference exceeds a predefined threshold, which is set to 10 pixels per frame.

\subsection{Object Softness Measurement}

\begin{figure}
    \centering
    \vspace{5pt}
    \includegraphics[width=\linewidth]{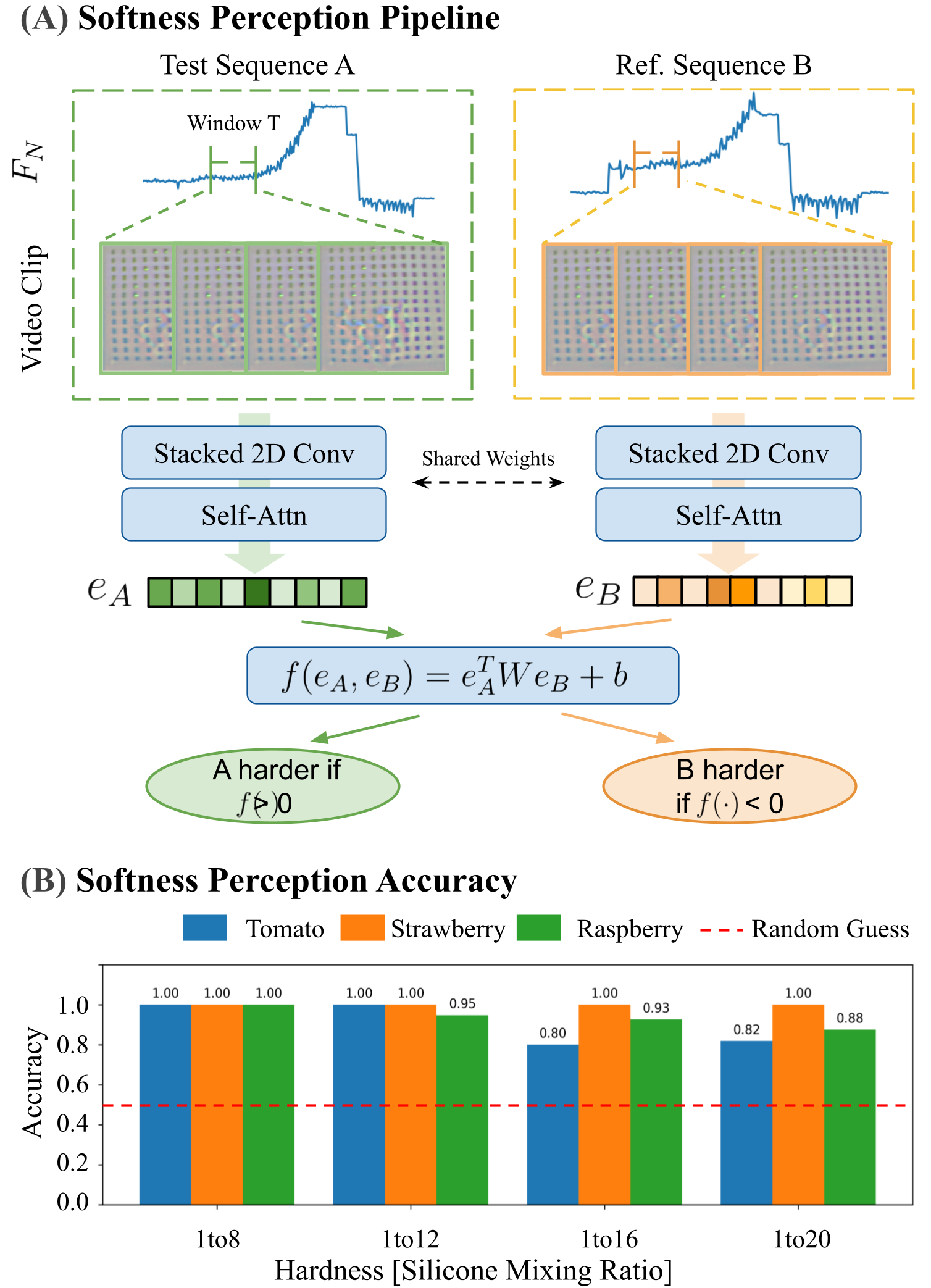}

    \caption{\textbf{(A)} Softness Prediction Pipeline. We randomly segment tactile image sequences and the corresponding normal force signals from the pressing process of the same fruit, and extract embeddings using a shared backbone. A lightweight ranking head then predicts whether the test object is harder or softer than the reference. \textbf{(B)} Softness perception results. Each bar shows classification accuracy for a specific hardness within a fruit type (comparisons within the same fruit). The red dotted line marks random guess (50\%).
    }
    \label{fig:softness}
    \vspace{-20pt}
\end{figure}

Determining whether a fruit is ready to be picked requires estimating its softness as a proxy for ripeness. Our gripper enables this capability in-hand: during a gentle squeeze, it combines normal force from motor current with tactile imprints from the gels to infer softness online, allowing pre-pick screening without additional labor. Moreover, the method is fruit-agnostic, with a single model generalizing across varieties without retraining.

Previous methods for estimating softness generally fall into two classes: classification-based \cite{doi:10.1126/sciadv.adp0348} and regression-based \cite{10521971, Yuan_2017_hardness}. Classification-based methods assign the object to one of several predefined firmness/ripeness categories (e.g., unripe/ready/overripe). In contrast, regression-based methods predict a continuous-valued mechanical property, such as Shore hardness, elastic modulus or compliance, providing a calibrated measure of softness. 
Classification simplifies downstream decisions but depends on well-chosen class boundaries and cannot generalize without retraining for each fruit; regression yields finer resolution but requires accurate ground-truth measurements, which is almost impossible in agriculture. 

Inspired by the human picking process, where we simply set a threshold in mind for the right picking ripeness and compare all samples with it among the same kind of fruit, we frame the softness prediction problem in agriculture as a pairwise comparison. Given two short compression sequences of the same fruit, the model predicts whether the first is harder than the second. This formulation avoids the need for absolute calibration, supports cross-variety generalization, and can be extended into an absolute softness scale through aggregation if desired.

We employ a lightweight ranker that encodes each compression clip with a short stack of 2D frame encoders followed by a self-attention to capture temporal features. A real-time estimate of normal force is projected and concatenated with the visual embedding. Given embeddings $\mathbf{e}_A, \mathbf{e}_B \in \mathbb{R}^D$, an asymmetric bilinear comparator
$f(\mathbf{e}_A,\mathbf{e}_B)=\mathbf{e}_A^\top W \mathbf{e}_B + b$ with $W=-W^\top$
produces a logit $f$. We train with binary cross-entropy on pair labels, forming within-fruit pairs. At inference, we predict ``$A>B$'' when $f \ge 0$. We visualize the model structure in \Cref{fig:softness}A.

\begin{figure*}
    \centering
        \vspace{5pt}
        \includegraphics[width=\linewidth]{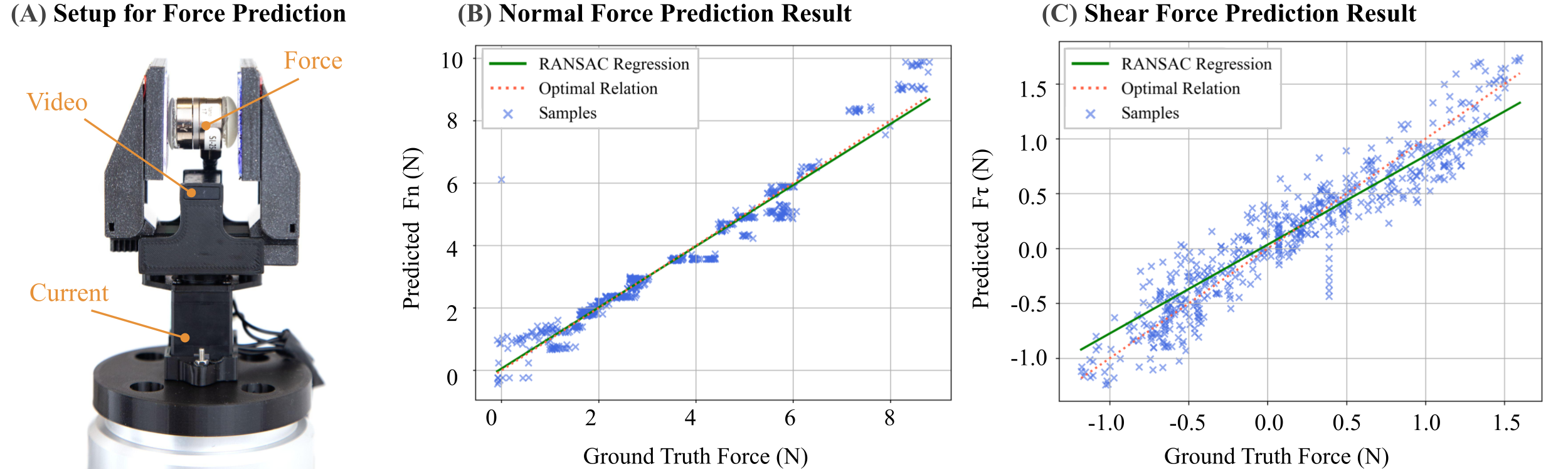}
    \caption{
    \textbf{Setup and Results for Force Prediction.} 
    \textbf{(A)} Experimental setup with the gripper mounted on an ATI Nano Force/Torque sensor, which measures the ground-truth force. 
    \textbf{(B, C)} Results of force estimation. 
    The \textbf{middle} panel compares normal force ($F_n$), while the \textbf{right} panel compares shear force ($F_{\tau}$). 
    For clarity, the magnitude of the shear force is used for visualization. 
    The solid green line indicates the RANSAC fit, and the dotted red line denotes the identity line ($y{=}x$).
    }

    \label{fig:force}
    \vspace{-15pt}
\end{figure*}

\section{Experiments and Results}



\subsection{Geometry Reconstruction}
\Cref{fig:calibration}B shows the output of both fingers and the reconstructed height map of the contact shape. Although the gripper is opened to different widths, our calibration model remains agnostic to the opening size and accurately estimates the corresponding normal maps.

To quantitatively assess sensor performance, we 3D-printed small objects with known geometries and compared the reconstructed height maps to ground truth. 
Compared to prior GelSight-style sensors that achieve sub-millimeter reconstruction accuracy under controlled indentation (e.g., \cite{gelslim1, mirzaee2025gelbelt},), our reconstruction error is comparable while using a single shared camera and a form factor suitable for in-field harvesting. Specifically, using a six-sided pyramid (10 mm diameter, 2 mm height) to indent the surface at multiple locations, we achieve an MSE of 0.201 mm², demonstrating that our sensor reconstructs contact features with high accuracy.

\subsection{Grasping Force Estimation}
For force estimation experiments, we mounted an ATI Nano17 force/torque sensor and grasped it with the proposed gripper at various locations, as shown in \Cref{fig:force}A.  We gradually closed the gripper while recording the force measurements and the motor current, resulting in approximately 10{,}000 data points collected across 20 trials using six different indenter shapes. 

For normal force estimation, the model predicts force from motor current with strong agreement to ground truth, achieving \(R^2 = 0.951\), MAE = 0.267~N and a mean absolute percentage error (MAPE) of 3.04\%. For the more challenging shear force estimation, the model attains \(R^2 = 0.903\), MAE = 0.149~N and MAPE = 9.37\%. Despite increased contact variability, both models demonstrate consistent performance across repeated trials and indenter geometries. \Cref{fig:force} illustrates the relationship between the measured and predicted forces in both directions. Under quasi-static grasping conditions similar to those used in prior agricultural force estimation studies~\cite{zhang2023flexible, rajendran2024enablingtactilefeedbackrobotic}, our method achieves comparable normal-force accuracy while additionally providing shear-force estimation within a compact, harvest-ready gripper.

\begin{figure}
    \centering
    \vspace{-10pt}    \includegraphics[width=\linewidth]{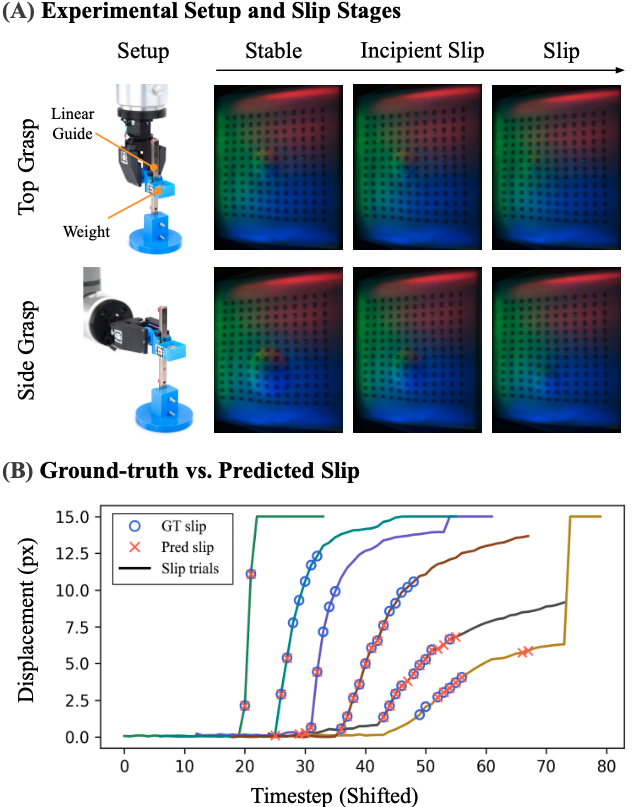}

    \caption{\textbf{Setup and results of slip detection.}  \textbf{(A)} Two grasp positions from the experiment, illustrating different slip stages (stable → incipient slip → full slip). \textbf{(B)} Time‐series comparison of ground‐truth and predicted slip events. The y-axis shows the AprilTag displacement from the initial pose, which we use as the ground-truth slip signal. Predictions are overlaid for better visualization.}
    \label{fig:slip}

    \vspace{-18pt}
\end{figure}

\subsection{Slip Detection}

We evaluated the slip detection system on a low-friction linear rail with adjustable loading, as shown in top row in \Cref{fig:slip}.
The gripper was tested in two grasping poses (top and side) to simulate diverse harvesting configurations, where the end-effector’s approach direction may be constrained by the environment. To modulate the sliding speed, we used three different loads: 10, 20, and 50 grams. Each slip detection condition was evaluated over approximately 200 frames per trial, with all load and pose combinations repeated twice. We use the AprilTag’s displacement from its starting position as the ground-truth indicator of slip.

The results are summarized in \Cref{fig:slip}, which shows sliding demonstrations and frame-by-frame prediction outputs for different trials. We adopt this frame-by-frame approach so it can be integrated directly with the controller for applications such as slip recovery. Quantitatively, the classifier attains a precision of \(0.725\), recall of \(0.661\), and F1 of \(0.692\). On correctly detected trials, the predicted slip typically precedes the first observable relative motion by 0.11 $\mathrm{s}$ on average, indicating preferential early reporting and enabling pre-emptive intervention. Compared to slip detection methods based on ultrasonic or high-frequency electrical sensing~\cite{agriculture12111802,LIU2024108904}, our approach trades absolute detection accuracy for tight integration with force estimation and closed-loop grasp control under realistic harvesting conditions.

\begin{figure*}
    \centering
        \vspace{5pt}\includegraphics[width=\linewidth]{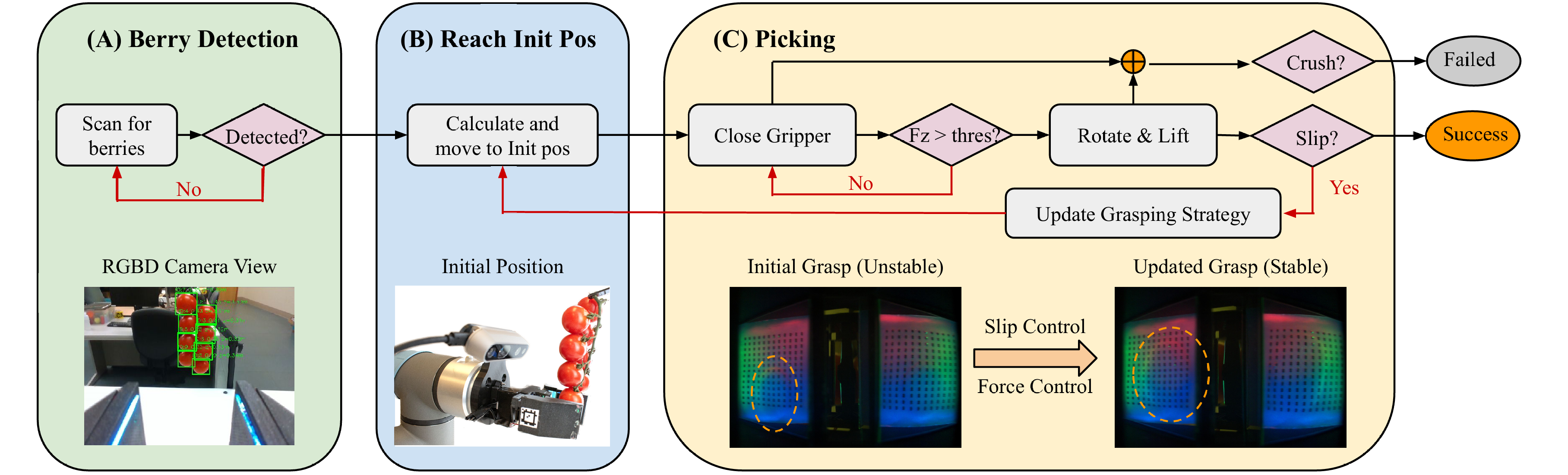}
    \caption{Flow diagram of harvest experiment, including berry detection, initial pose calculation, and picking phases. \textbf{(A)} RGBD camera to scan the area and identify berries. \textbf{(B)} Align and move the manipulator to initial pose. \textbf{(C)} The gripper executes the picking action according to the chosen control strategy.}
    \label{fig:system}
    \vspace{-15pt}
\end{figure*}

\subsection{Softness Measurement}
To systematically collect grasping data across different textures and hardness levels, we fabricated silicone replicas of three types of fruit: strawberry, raspberry, and cherry tomato using custom molds. For each fruit, we prepared four hardness variants by mixing two-part silicones (XP-565, Silicone Inc.) in different ratios (1: 8, 1:12, 1:16, and 1:20 for parts A and B). This results in hardness values of 68.4, 64.8, 51.4, and 42.2 on the Shore 00 scale. For each object, we conducted ten grasping trials at varied positions while recording video data of the interactions. Normal force was estimated using the proposed linear model, and the resulting signals were used to train the pairwise classification network.

\Cref{fig:softness}B illustrates the model’s performance on the test sequences. Each bar shows classification accuracy for a specific hardness level within a fruit type, with comparisons restricted to fruits of the same kind. In general, fruits with complex surface textures (e.g., strawberry, raspberry) yield higher accuracy than textureless fruits (e.g., cherry tomato), as the richer contact patterns provide more informative tactile signals. This suggests that surface geometry plays an important role in facilitating reliable softness estimation.  Aggregated across all fruits, the model achieves \(\sim\!94.7\%\) validation accuracy.

\subsection{Robot System For Harvesting}

We integrate the proposed FruitTouch gripper into a fully autonomous robotic system to validate its performance in cherry tomato and strawberry harvesting. As illustrated in \Cref{fig:system}, the system uses a state machine for motion planning, while force estimation and slip detection run in a unified perception–control loop at approximately 15 Hz, improving grasp stability and responsiveness during picking. The gripper is mounted on a UR5e robotic arm. Cherry tomatoes with intact vines are suspended by fixing the vine tips to a pole within the robot’s reachable workspace, preserving their natural clustered structure. Strawberries are similarly suspended using their natural stems to emulate typical hanging harvesting configurations. Using a RealSense D435 camera, a YOLOv5 detector first localizes the fruits in the world frame, after which the robot end effector moves to an appropriate grasping pose.

To highlight the importance of enhanced sensing at the end effector, we conducted ablation experiments to compare three control strategies: open-loop, slip control, and slip+force control. In the open-loop setting, once the fruit’s position and diameter were detected, the gripper closed to a fixed width (fruit diameter minus 2 mm) to attempt the pick. In the slip control strategy, slip events were continuously monitored. Upon detecting slip, the gripper closed an additional 2 mm before retrying. In slip+force control, retries were guided by force feedback: for cherry tomatoes, the initial normal force was set to 1.2 N and increased in 0.3 N increments, while for strawberries with stiffer stems, the initial force was 2 N and increased by 1 N per attempt. 
Each strategy was evaluated on 12 cherry tomatoes and 8 strawberries, with a grasp considered successful if the fruit was detached without bruising or sliding onto the table. As shown in \Cref{tab:strategy-compare}, slip+force control achieves the highest picking success with the lowest force variability, at the cost of slightly more retries, demonstrating reliable and gentle fruit harvesting.




\begin{table}[t]
\centering
\caption{Real Fruit Harvesting Results (Lab Setting)}
\vspace{-8pt}
\label{tab:strategy-compare}
\setlength{\tabcolsep}{4pt}

\begin{threeparttable}

\begin{tabular}{lccccc}
\toprule
\textbf{Fruit / Strategy} 
& \textbf{Succ. (\%)} 
& \textbf{Attempts} 
& \textbf{$F_z$ Mean / Var ($N$, $N^2$)} \\
\midrule

\multicolumn{4}{l}{\textit{Cherry tomato}} \\
\midrule
Open-loop          
& $58.3$  
& $-$     
& $-$ \\
Slip control       
& $100$   
& $1.167$ 
& $1.571 / 0.506$ \\
Slip+Force control 
& $100$   
& $1.417$ 
& $1.557 / 0.062$ \\

\midrule
\multicolumn{4}{l}{\textit{Strawberry}} \\
\midrule
Open-loop          
& $12.5$  
& $-$     
& $-$ \\
Slip control       
& $37.5$   
& $1.625$ 
& $2.648/ 0.495$ \\
Slip+Force control 
& $75$   
& $2.750$ 
& $3.089 / 0.231$ \\

\bottomrule
\end{tabular}

\begin{tablenotes}
\footnotesize
\item Success is defined as detaching the fruit without damage. 
\end{tablenotes}

\end{threeparttable}
\vspace{-15pt}
\end{table}

\section{Conclusion}
We presented FruitTouch, a compact, low-cost, customizable, sensorized gripper that brings high-resolution tactile perception, including geometry reconstruction, force estimation, slip detection, and in-hand softness assessment into a practical harvesting system. 
In controlled evaluations, the gripper achieved accurate force estimation, timely slip detection, and fruit-agnostic pairwise softness ranking, while harvesting ablations confirmed the benefits of tactile feedback. In particular, tactile-informed closed-loop control improves grasp robustness while reducing excessive contact forces, enabling gentle and adaptive manipulation. Together, these results address key challenges in automated harvesting by supporting damage-free picking of fragile fruits, maintaining cost-efficient and scalable design, and enabling in-hand sensing for fruit quality assessment.




\addtolength{\textheight}{-0cm}   
\bibliographystyle{IEEEtran}
\bibliography{ref}

\newpage

\end{document}